\documentclass[runningheads]{llncs}

\usepackage[T1]{fontenc}
\usepackage{algorithm}
\usepackage{algorithmic}
\usepackage{amsmath,amssymb,amsfonts}
\usepackage{bbm}
\usepackage{booktabs}
\usepackage{float}
\usepackage{graphicx}
\usepackage[hidelinks]{hyperref}
\usepackage{textcomp}
\usepackage{subcaption}
\usepackage{url}
\usepackage{xcolor}
\usepackage{xspace}

\newcommand{\alg}{\textsc{KANG}\xspace}

\begin{document}

\title{Kolmogorov--Arnold Graph Neural Networks}
\titlerunning{Kolmogorov--Arnold Graph Neural Networks}

\author{Gianluca De Carlo\inst{1} \and
Andrea Mastropietro\inst{2,3} \and
Aris Anagnostopoulos\inst{1}}
\authorrunning{G. De Carlo et al.}
\institute{Department of Computer, Control and Management Engineering,\\
Sapienza University of Rome, Italy \\
\and
Department of Life Science Informatics and Data Science, B-IT, LIMES Program Unit Chemical Biology and Medicinal Chemistry, Rheinische Friedrich-Wilhelms-Universität, Bonn, Germany
\and
Lamarr Institute for Machine Learning and Artificial Intelligence, Bonn, Germany\\
\email{\{decarlo,aris\}@diag.uniroma1.it}\\
\email{mastropietro@bit.uni-bonn.de}}
\maketitle

\begin{abstract}
Kolmogorov--Arnold Networks (KANs) recently emerged as a powerful
alternative to traditional multilayer perceptrons, providing enhanced
generalisation and intrinsic interpretability through learnable
spline-based activation functions. Motivated by the need for powerful
and transparent graph neural networks (GNNs), we propose the
Kolmogorov--Arnold Network for Graphs (\alg), a novel GNN architecture
that integrates KANs into the message-passing framework. Specifically,
we demonstrate that a data-aligned initialisation of the splines,
coupled with increased flexibility during training, significantly
enhances model performance. Experiments on benchmark datasets
for node classification, link prediction, and graph classification tasks
show that \alg consistently outperforms established GNN architectures.
Additionally, we analyse the robustness of \alg to oversmoothing,
confirming its ability to maintain expressive node representations even
in deeper network architectures. Our findings position \alg as a
promising alternative to traditional GNNs, combining superior
performance with a foundational step towards transparent GNNs.
Code is available at \url{https://anonymous.4open.science/r/KANGnn-1B07}.
\end{abstract}

\section{Introduction}
Traditionally, multilayer perceptrons (MLPs)~\cite{haykin1998neural}
have served as the foundational building blocks of neural networks.
Recently, however, Kolmogorov--Arnold Networks (KANs)~\cite{liu2024kan}
have emerged as a promising alternative. KANs, inspired by the Kolmogorov--Arnold representation theorem~\cite{Kolmogorov1956,Kolmogorov1957}, provide an alternative to traditional neural architectures by leveraging learnable spline-based activations, improving both generalization and interpretability.

In parallel, Graph Neural Networks (GNNs)~\cite{scarselli2008graph} have emerged as powerful models for learning from graph-structured data. By leveraging connectivity patterns, GNNs generate topology-aware embeddings. Notable architectures include Graph Convolutional Networks (GCNs)~\cite{semikipf2017}, which extend convolution operations to graph domains; GraphSAGE~\cite{hamilton2017inductive}, which inductively aggregates neighborhood information; Graph Attention Networks (GATs)~\cite{velickovic2018}, incorporating attention to identify relevant neighbors; and Graph Isomorphism Networks (GINs)~\cite{DBLP:conf/iclr/XuHLJ19}, achieving discriminative power akin to the Weisfeiler–Lehman test~\cite{weisfeiler1968reduction}. Extensions such as Relational GCNs (RGCNs) address heterogeneous graphs by explicitly modeling relational information, and Variational Graph Autoencoders (VGAEs)~\cite{DBLP:journals/corr/KipfW16a} facilitate latent-space representation learning. Collectively, these methods have significantly advanced applications in social networks, biology, and knowledge graphs~\cite{wu2020comprehensive}.

In this work, the strengths of KANs and the expressive power of GNNs are
bridged by extending the Kolmogorov--Arnold representation theorem to
graph-based neural networks. We introduce the Kolmogorov--Arnold Network
for Graphs (\alg), a unified framework designed to leverage transparent
spline-based activation functions within a graph neural architecture.
Specifically, our contributions include:

\begin{itemize}
\item \textbf{Novel GNN Architecture:} We introduce \alg, a novel GNN architecture using spline-based activations to enhance transparency and generalization, while preserving the expressive power of MLPs and message-passing GNNs. Additionally, we propose KAN Distribution (KAND), a variant of KAN with data-driven initialisation and learnable spline control points.

\item \textbf{Superior Performance:} \alg consistently outperforms established GNNs on standard benchmarks (Cora, PubMed, CiteSeer, MUTAG, PROTEINS) for node classification, link prediction, and graph classification.

\item \textbf{Comprehensive Analysis:} We examine \alg's key properties, including the impact of data-aligned initialisation, robustness to oversmoothing, and computational scalability.
\end{itemize}

The remainder of the paper is organized as follows. We first review related work and present essential theoretical background. Next, we introduce the proposed \alg architecture and evaluate it experimentally, demonstrating its superior performance over existing GNNs. We conclude with a discussion on implications and directions for future research.

\section{Related Work}
Following the foundational work of Liu et al.~\cite{liu2024kan}, a surge
of research has extended KANs across diverse
domains~\cite{somvanshi2024survey}. For example, KAN variants have been
developed for time-series forecasting~\cite{genet2024tkan},
self-supervised mesh generation~\cite{zhang2025meshkinn}, and
specialized tasks such as wavelet-based KANs for signal
processing~\cite{meshir2025study}. Additionally, KANs have been
successfully integrated into transformers~\cite{vaswani2017attention},
leading to performance improvements in various
applications~\cite{yang2024kolmogorov}. The continuous evolution of
KAN-based models across different domains underscores their potential
benefits. Overall, KANs are emerging as a flexible framework that
preserves universal approximation capabilities while enhancing
interpretability in deep learning models.

GNNs have become the de facto approach for learning from
graph-structured data. GNNs can be categorized into three main spatial
flavours~\cite{velivckovic2023everything}, namely \textbf{convolutional
GNNs}, which aggregate node embeddings with fixed learnable
parameters~\cite{semikipf2017}, \textbf{attentional GNNs}, which assign
adaptive neighbor-specific weights to refine
aggregation~\cite{velickovic2018,brody2021attentive}, and
\textbf{message-passing GNNs}, which extend this framework by utilizing
a learnable function to compute node representations based on
neighborhood
interactions~\cite{battaglia2016interaction,gilmer2017neural}. Notably,
the integration of KANs with graph-structured data has recently gained
significant interest, and for each of these paradigms, corresponding
KAN-based GNN architectures have been explored. For instance, Kiamari et
al.~\cite{kiamari2024gkan} proposed two KAN-based architectures: One
where node embeddings are aggregated before applying spline-based KAN
layers, and another where KAN layers are applied prior to aggregation.
Their experiments, however, were limited to a reduced subset of features
from the Cora dataset (only 200 out of 1433). Similarly, Bresson et
al.~\cite{bresson2024kagnns} introduced two GNN variants that
incorporate KAN layers into node representation updates---KAGIN
(inspired by GIN) and KAGCN (based on GCN). Additionally, Ahmed et
al.~\cite{ahmed24graphkan} applied a KAN-based GNN architecture to
molecular graph learning, specifically for predicting protein--ligand
affinity.

While these works represent significant progress in integrating KANs
with GNNs, several open challenges remain. For example, the limited
feature subset used~\cite{kiamari2024gkan} might not fully showcase the
potential of KAN-based models. Other studies~\cite{bresson2024kagnns} neither thoroughly examine key architectural decisions nor explore critical issues like oversmoothing robustness in detail. Furthermore,
certain implementations~\cite{zhang2024graphkan} lack comprehensive
experimental validation across diverse datasets, potentially limiting
the generalizability and robustness of their conclusions.

Our work directly addresses these limitations by introducing a more
comprehensive and rigorously validated KAN-based GNN architecture.
Specifically, we extend the Kolmogorov--Arnold representation theorem to
graph neural networks and perform a detailed analysis of model
characteristics, transparency, and performance across multiple tasks and
datasets.

\section{Background}\label{sec:background}
In this section, we discuss the foundational concepts necessary to understand the proposed \alg model. We begin by introducing KANs and their key properties, followed by a brief overview of message passing neural networks (MPNNs). We then discuss the computational efficiency of spline functions, which are a fundamental component of the \alg model, and explore how their computation can be optimized.

\subsection{Kolmogorov--Arnold Networks}
The Kolmogorov--Arnold theorem states that any multivariate continuous function on a bounded domain can be rewritten using a finite composition of continuous univariate functions and summation operations. Given $\mathbf{x} \in \mathbb{R}^n$ and a function $f: \left[0,1\right]^n \rightarrow \mathbb{R}$, it is possible to express it as:

\begin{equation}\label{eq:2-layer-kan}
    f(\mathbf{x}) = \sum_{q=1}^{2n+1} \Phi_q\left(\sum_{p=1}^{n} \phi_{q,p}(x_p)\right),
\end{equation}
where $\phi_{q,p}: \left[0,1\right] \rightarrow \mathbb{R}$ are continuous functions applied to individual input features, and $\Phi_q: \mathbb{R} \rightarrow \mathbb{R}$ are additional continuous functions applied to the aggregated results. 

Liu et al.~\cite{liu2024kan} extended Equation~\ref{eq:2-layer-kan} by parameterizing the univariate functions $\phi$ as B-spline curves with learnable coefficients. However, Equation~\ref{eq:2-layer-kan} describes only a two-layer KAN, which may lack sufficient expressiveness for complex tasks. This limitation can be mitigated by introducing deeper architectures, similar to MLPs:

\begin{equation}
    \label{eq:kan_stack}
    \text{KAN}(\mathbf{x}) = (\Phi_{L-1} \circ \phi_{L-1} \circ \ldots \circ \Phi_1 \circ \phi_1)(\mathbf{x}),
\end{equation}

where $L$ is the number of layers. Each function $\phi$ can be expressed as a combination of a standard activation function and a spline-based transformation:

\begin{equation}
    \phi(\mathbf{x}) = w_b b(\mathbf{x}) + w_s \mathrm{spline}(\mathbf{x}),
\end{equation}
where $b(\mathbf{x})$ is a basis function (commonly a $\mathrm{SiLU}$ activation), $\mathrm{spline}(\mathbf{x})$ is a B-spline function, and $w_b, w_s$ are learnable weights. The spline function is formulated as a sum of B-spline basis functions $B_i(\mathbf{x})$, each with a learnable coefficient $c_i$:

\begin{equation}
    \mathrm{spline}(\mathbf{x}) = \sum_{i=1}^{n} c_i B_i(\mathbf{x}).
\end{equation}

\subsection{Activation Functions}\label{sec:back-splines}
\subsubsection{B-splines}
B-splines (basis splines) are piecewise polynomial functions widely used in numerical analysis and computer-aided geometric design~\cite{cohen1980discrete}. They are defined over a sequence of knots (or control points) that partition the domain into intervals, within which polynomial segments of degree \( p \) are defined. A B-spline of degree \( p \) ensures continuity up to order \( p-1 \) at the knots.

The recursive definition of B-spline basis functions \( B_{i,p}(t) \) follows the Cox–de Boor formula:

\[
\begin{aligned}
B_{i,0}(t) &= 
\begin{cases}
1 & \text{if } t_i \leq t < t_{i+1}, \\
0 & \text{otherwise},
\end{cases} \\
B_{i,p}(t) &= \frac{t - t_i}{t_{i+p} - t_i} B_{i,p-1}(t) + \frac{t_{i+p+1} - t}{t_{i+p+1} - t_{i+1}} B_{i+1,p-1}(t),
\end{aligned}
\]
where \( t_i \) are the knot positions.

B-splines exhibit several desirable properties:

\begin{itemize}
    \item \textbf{Local Support}: Each basis function \( B_{i,p}(t) \) is nonzero only within a limited interval \( [t_i, t_{i+p+1}) \), enabling localized control over the spline shape.
    \item \textbf{Partition of Unity}: The basis functions satisfy \( \sum_i B_{i,p}(t) = 1 \), ensuring that the spline remains within the convex hull of its control points.
    \item \textbf{Smoothness}: B-splines maintain \( C^{p-1} \)-continuity, ensuring smooth transitions between polynomial segments.
\end{itemize}

Although B-splines provide strong theoretical guarantees and flexibility, their recursive evaluation can be computationally expensive, especially for high-degree splines or a large number of control points. Additionally, their recursive nature limits efficient parallelization on modern GPUs~\cite{yang2024kolmogorov}.

\subsubsection{Radial Basis Functions}\label{sec:rbf}
Radial Basis Functions (RBFs) are a class of functions with radial symmetry, meaning their output depends only on the distance from a central point. Due to their universal approximation properties and computational efficiency, RBFs have gained popularity as alternatives to splines~\cite{li2024kolmogorov} in function approximation tasks. In an RBF network, the output is computed as:

\begin{equation}\label{eq:rbf}
y(\mathbf{x}) = \sum_{i=1}^{N} w_i\, \varphi\bigl(\|\mathbf{x} - \mathbf{c}_i\|\bigr),
\end{equation}
where \( w_i \) are learnable weights, \( \mathbf{c}_i \) are the RBF centers, and \( \varphi \) is a chosen radial basis function (e.g., Gaussian or multiquadric kernel). One commonly used function is the Gaussian RBF:

\begin{equation}
\varphi(r) = e^{-\gamma r^2},
\end{equation}
where \( \gamma \) controls the width of the kernel and \( r \) is the Euclidean distance. RBFs are computationally more efficient than high-degree splines and have been successfully applied in support vector machines (SVMs) and kernel methods~\cite{cortes1995support}.

\subsection{Message Passing Neural Networks}
MPNNs provide a general framework for learning node representations through iterative information exchange and aggregation among neighboring nodes. At each layer, a node \( u \) updates its representation by aggregating messages from its neighbors:

\begin{equation}
    \mathbf{h}_u = \phi\left(\mathbf{x}_u, \bigoplus_{v \in \mathcal{N}(u)} \psi(\mathbf{x}_u, \mathbf{x}_v)\right),
\end{equation}
where \( \mathbf{x}_u, \mathbf{x}_v \in \mathbb{R}^d \) are feature vectors of nodes \( u \) and \( v \), \( \mathcal{N}(u) \) is the set of neighbors of \( u \), and $\bigoplus$ denotes a differentiable, permutation invariant function (e.g., sum, mean, max). The function \( \psi(\mathbf{x}_u, \mathbf{x}_v) \) computes a message from receiving and sending nodes, for example $\psi(\mathbf{x}_u, \mathbf{x}_v) = \text{MLP}(\mathbf{x}_u||\mathbf{x}_v))$. Finally, the update function \( \phi \) integrates the aggregated neighborhood information with the node's own features, often modeled as an MLP, for example: $ \phi(\mathbf{x}_u, \mathbf{m}_u) = \text{MLP}(\mathbf{x}_u \| \mathbf{m}_u])$ where \( \mathbf{m}_u = \bigoplus_{v \in \mathcal{N}(u)} \psi(\mathbf{x}_u, \mathbf{x}_v) \).

\section{Methodology}\label{sec:methodology}
\begin{figure}[ht]
    \centering
    \includegraphics[width=\textwidth]{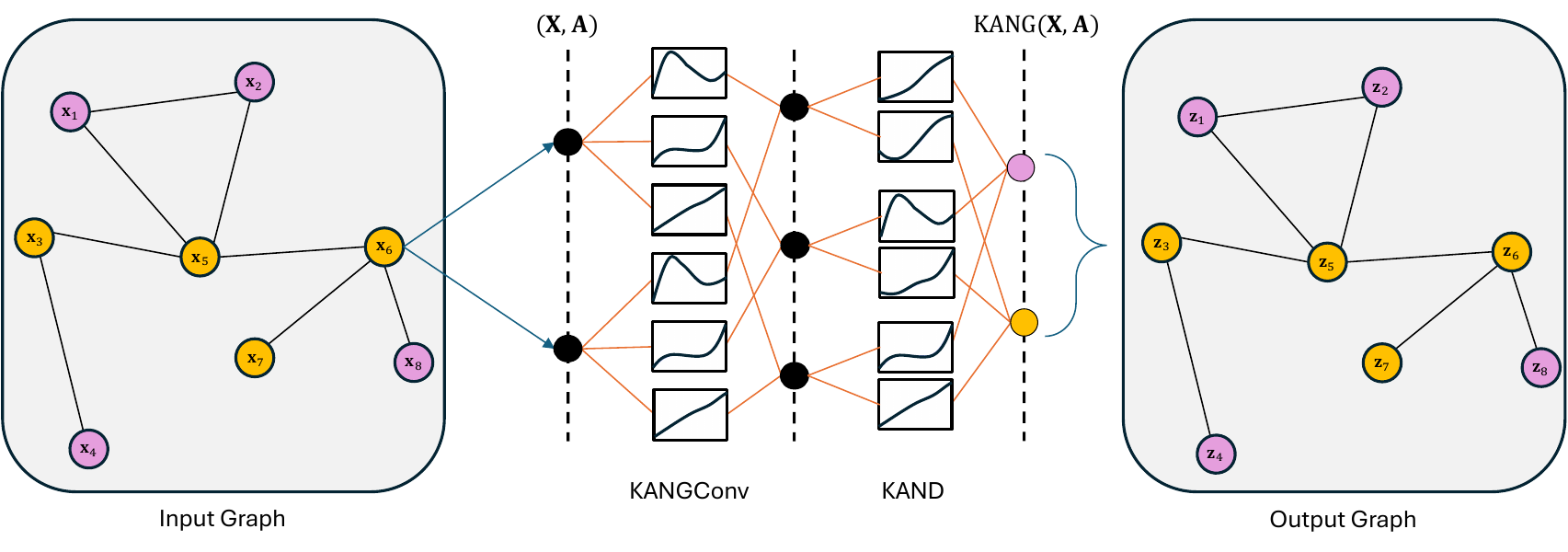}
    \caption{Illustration of the \alg architecture. The input graph is processed through multiple
      KANG convolutional layers, where spline-based transformations
      dynamically aggregate and update node embeddings. A KAND layer refines the final node embeddings for specific downstream tasks.}
    \label{fig:kang}
\end{figure}

In this section, we present \alg, an innovative GNN architecture that incorporates the KAN paradigm within the message-passing framework (Figure~\ref{fig:kang}). To accomplish this, we introduce KAND, a new KAN-based model featuring learnable control points with initialisation aligned to the input distribution.
This approach enables expressive and transparent node representation learning by leveraging spline-based transformations within graph convolutional layers. Implementation details can be found in Appendix~\ref{A:impl}.

\subsection{KAND}
\begin{figure}[ht]
    \centering
    \begin{subfigure}[b]{0.49\textwidth}
        \centering
        \includegraphics[width=\textwidth]{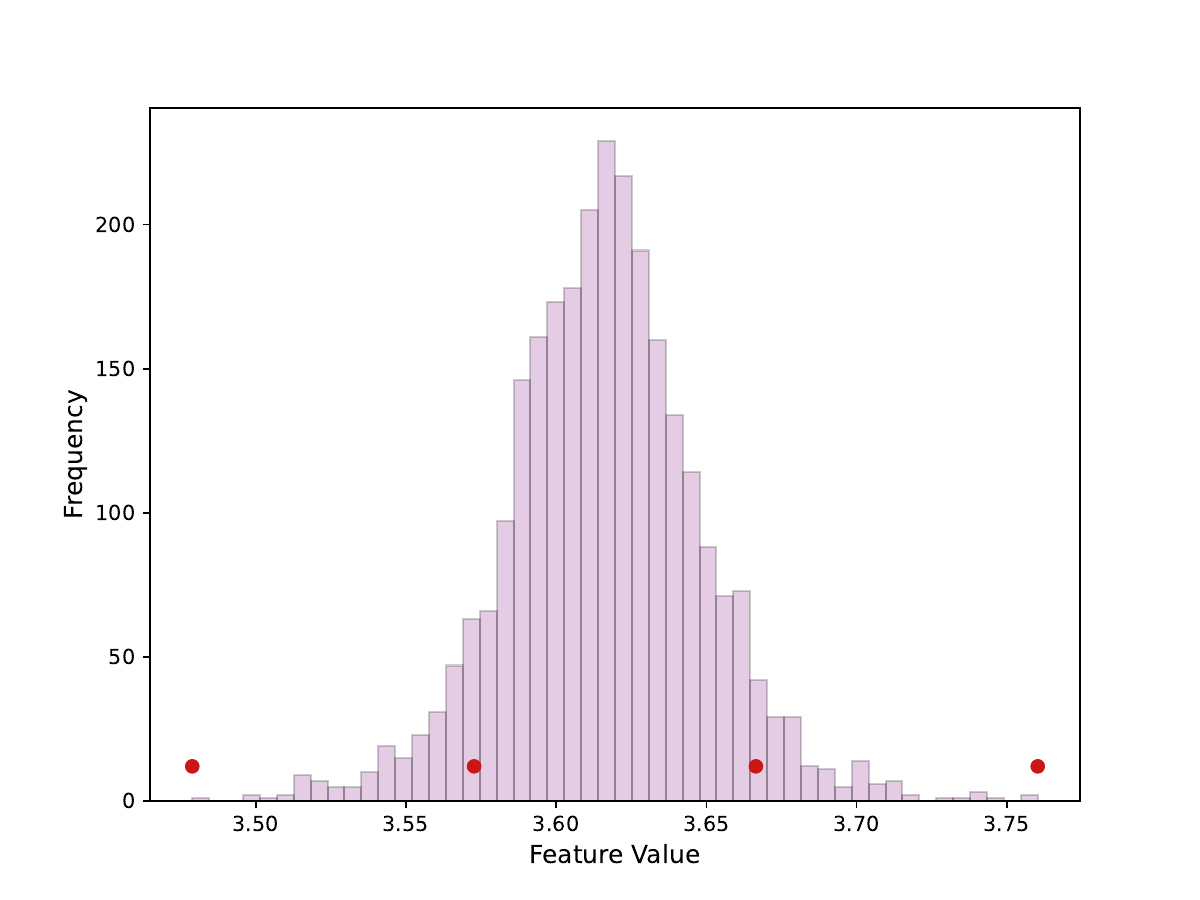}
        \caption{Feature distribution without LN.}
        \label{fig:not_norm}
    \end{subfigure}
    \begin{subfigure}[b]{0.49\textwidth}
        \centering
        \includegraphics[width=\textwidth]{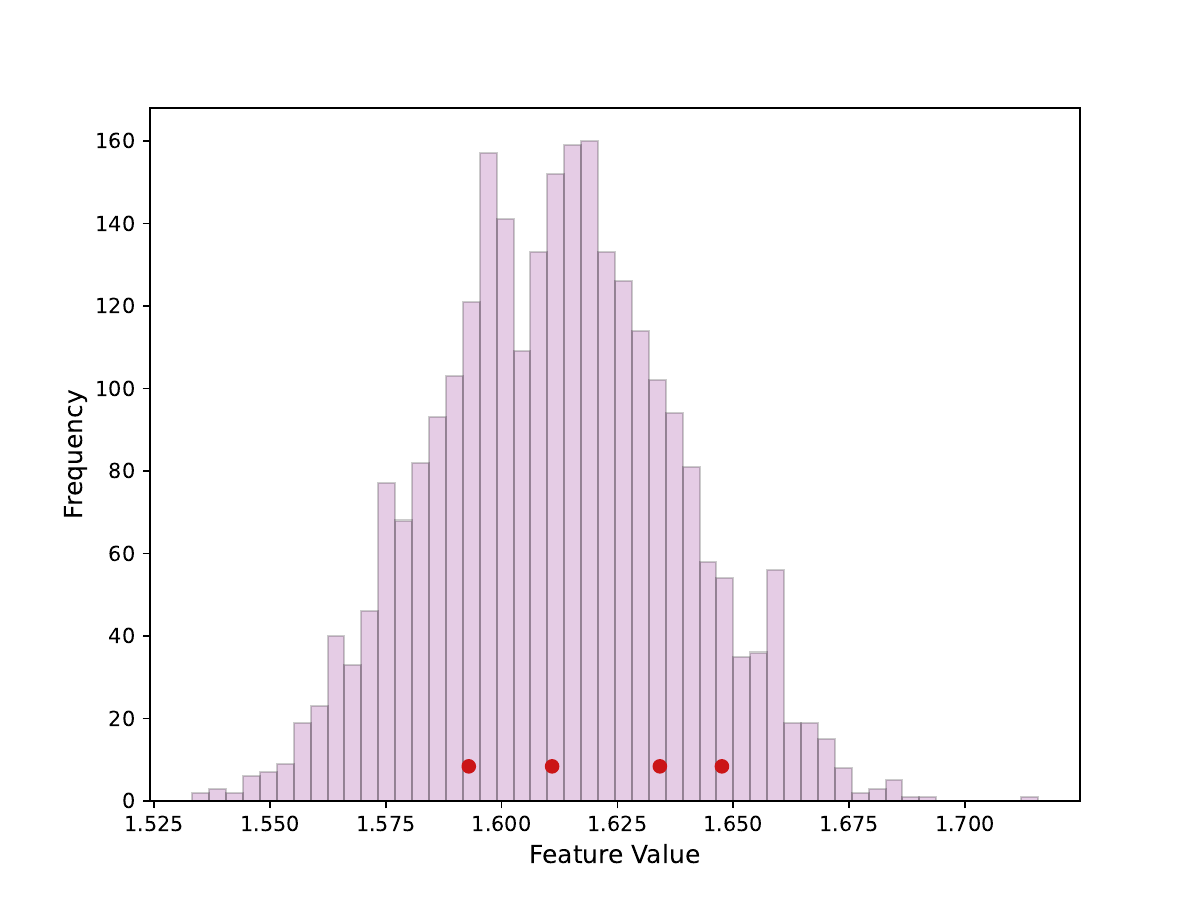}
        \caption{Feature distribution with LN.}
        \label{fig:norm}
    \end{subfigure}
    \caption{Comparison of feature distributions after the first KANG layer for a selected feature dimension.
    \textbf{Left:} although the distribution visually appears Gaussian-like, the evenly spaced spline control points (red dots) do not accurately reflect the underlying density of the feature values. This rigid placement may limit the spline's flexibility in adapting to the regions of highest feature density.
    \textbf{Right:} applying LN produces a standardized and more compact Gaussian-like distribution. The spline control points sampled according to a Gaussian distribution, align closely with the density of feature values, thus improving representational flexibility and expressiveness of the spline activations within KANG layers.}
    \label{fig:norm-all}
\end{figure}

In KANs, the features $x_p$ of an input $\boldsymbol{x}\in\mathbb{R}^n$ are rescaled to a region $[a,b]$, and the control points $\{t_i\}_{i=1,\dots,K}$ are evenly spaced in $[a,b]$. However, this might not be the optimal distribution for the control points, as they might over-represent or under-represent certain features of the input (Figure \ref{fig:not_norm}).

To overcome this limitation, we introduce the KAN Distribution (KAND), a method whereby we impose constraints on the distribution of input features and subsequently sample the control points accordingly. Specifically, we employ layer normalization (LN) to ensure that the input features conform to a distribution characterized by zero mean and unit standard deviation.

While this methodology ensures a proportional representation of the features' domain to learn the splines, it does not guarantee its optimality. Therefore, unlike traditional KANs, we also learn the displacement of each $t_i$ from its initial location. We discuss the impact of this addition in section~\ref{sec:ablation_control}.

While spline-based functions effectively capture complex nonlinearities, their recursive evaluation through the Cox–de Boor algorithm is computationally demanding and inadequately optimized for GPU parallelization (see Section~\ref{sec:back-splines}). To mitigate these computational drawbacks, we substitute spline activations with RBFs, exploiting their closed-form expressions to achieve improved computational efficiency without sacrificing generalization performance \cite{li2024kolmogorov}. Therefore, KANDs' architecture takes the form
\begin{equation}
    \text{KAND}(\mathbf{x}) = (\text{LN}\circ \Phi_{L} \circ \tilde{\phi}_{L} \circ \ldots \circ \text{LN} \circ \Phi_1 \circ \tilde{\phi}_1)(\mathbf{x}),
\end{equation}
where $\tilde{\phi}$ are RBF kernels with learnable control points and $\Phi$ are linear layers. We empirically analyze the impact of substituting B-splines with RBFs in Section~\ref{sec:ablation-rbf}. 

\subsection{KANG Convolution}
The core component of our model is the \textbf{KANGConv} operator, which generalizes graph convolution by incorporating spline-based activation functions inspired by the Kolmogorov--Arnold framework. Unlike conventional GNN layers, which rely on static nonlinearities (e.g. ReLU), KANGConv dynamically adapts its activation functions to the data.

Let \( \mathbf{h}_u^{(l)} \in \mathbb{R}^{d} \) denote the embedding of node \( u \) at layer \( l \), where \( d \) is the feature dimension. The update rule for node embeddings is given by:

\begin{equation}
    \mathbf{h}_u^{(l+1)} = \text{KAND} \left( \mathbf{h}_u^{(l)} + \sum_{v \in \mathcal{N}(u)} \psi_{\text{KAND}}(\mathbf{h}_u^{(l)}, \mathbf{h}_v^{(l)}) \right),
\end{equation}

where $\psi_{\text{KAND}}(\mathbf{h}_u^{(l)}, \mathbf{h}_v^{(l)})=\text{KAND}([\mathbf{h}_u^{(l)}||\mathbf{h}_v^{(l)}])$, and \( \mathcal{N}(u) \) denotes the neighborhood of node \( u \). The function \( \psi_{\text{KAND}}(\mathbf{h}_u^{(l)}, \mathbf{h}_v^{(l)}) \) computes a learnable transformation of the node features, leveraging adaptive spline-based activations. This formulation extends traditional message-passing models by allowing the network to learn feature-dependent activation functions, rather than relying on a fixed nonlinearity.

The proposed architecture consists of multiple stacked KANGConv layers, each performing localized feature aggregation through learnable spline transformations. By capturing complex, data-dependent nonlinear interactions, KANG enables more expressive message passing while maintaining the theoretical transparency of spline-based functions.

The final node representations are projected through a KAND layer, whose role is to refine embeddings for downstream tasks such as node classification, link prediction, and graph classification.

\section{Experimental Evaluation}\label{sec:experiments}
In this section, we evaluate the proposed KANG architecture across three standard tasks in graph representation learning: node classification, link prediction, and graph classification. We systematically benchmark KANG against established GNN architectures, specifically GCN, GATv2, GraphSAGE, and GIN.

\subsection{Datasets}
Our evaluation encompasses widely recognized benchmark datasets detailed in Table~\ref{tab:datasets}. For node classification and link prediction tasks, we employ the public splits provided for the citation datasets Cora, CiteSeer, and PubMed. Conversely, for graph classification tasks, specifically on MUTAG and PROTEINS datasets, we employ stratified random splits, allocating 80\% of samples for training, 10\% for validation, and 10\% for testing, to ensure robust statistical evaluation. Dataset details can be found in Appendix~\ref{A:data}

\subsection{Experimental Setup}
Hyperparameter optimization across all evaluated models was performed via grid search, exploring learning rate \(\in \{5\times 10^{-4}, ..., 10^{-2}\}\), weight decay \(\in [10^{-4}, 5 \times 10^{-3}]\), hidden channels \(\in \{8, 16, 32, 64\}\), dropout rates in the range \([0, 0.6]\), and the number of layers $\in [2,3,4]$.

Given the unique characteristics of KANG, three additional hyperparameters were optimized, namely the number of spline control points (grid size ranging from 2 to 10), and minimum and maximum spline grid values (explored in ranges \([-20,0]\) and \([0,20]\), respectively). A sensitivity analysis of the minimum and maximum spline grid values is provided in the~\ref{A:grid}.

Training was conducted over at most 1000 epochs using the AdamW~\cite{loshchilov2017decoupled} optmizer, with an early stopping mechanism triggered if validation performance did not improve for 300 consecutive epochs. To ensure fair comparisons and reproducibility, each experimental run used a fixed random seed, ensuring consistency across all tested architectures. Reported metrics represent average results computed from 30 independent experimental repetitions.

\subsection{Results}
Table~\ref{tab:results} summarizes the empirical results\footnote{All experiments were carried out on a g5.xlarge AWS instance.}, indicating that KANG consistently achieves superior or competitive performance across all tasks and datasets evaluated. These findings confirm the effectiveness and versatility of our spline-based graph neural network architecture, achieving competitive performance in most tasks.
\begin{table}[h]
\centering
\caption{Performance comparison of KANG against baseline GNN architectures, reporting average test accuracy (\%) for node and graph classification and test AUC-ROC for link prediction tasks. Results are presented as the mean $\pm$ standard deviation over 30 runs.}
\label{tab:results}
\resizebox{\textwidth}{!}{
\begin{tabular}{lccccc}
\toprule
\textbf{Dataset} & \textbf{GCN} & \textbf{GATv2} & \textbf{GraphSAGE} & \textbf{GIN} & \textbf{KANG (Ours)} \\
\midrule
\multicolumn{6}{l}{\textbf{Node Classification}} \\
Cora & 79.9 $\pm$ 0.9 & 79.7 $\pm$ 1.2 & 77.9 $\pm$ 0.9 & 80.0 $\pm$ 1.1 & \textbf{82.0 $\pm$ 0.7} \\
PubMed & 77.5 $\pm$ 0.7 & 77.3 $\pm$ 0.7 & 76.3 $\pm$ 0.8 & 76.8 $\pm$ 0.7 & \textbf{78.0 $\pm$ 0.8} \\
CiteSeer & 67.5 $\pm$ 1.1 & 67.2 $\pm$ 1.5 & 66.2 $\pm$ 1.3 & 67.4 $\pm$ 1.5 & \textbf{70.4 $\pm$ 1.3} \\

\multicolumn{6}{l}{\textbf{Link Prediction}} \\
Cora & 82.0 $\pm$ 3.9 & 82.1 $\pm$ 9.5 & 86.7 $\pm$ 8.8 & \textbf{91.7 $\pm$ 0.9} & 91.4 $\pm$ 0.9 \\
PubMed & 86.5 $\pm$ 2.4 & 87.0 $\pm$ 2.3 & 92.6 $\pm$ 4.0 & 93.2 $\pm$ 0.5 & \textbf{95.2 $\pm$ 0.3} \\
CiteSeer & 81.6 $\pm$ 4.2 & 84.2 $\pm$ 1.5 & 91.3 $\pm$ 3.2 & 90.7 $\pm$ 1.3 & \textbf{92.4 $\pm$ 1.0} \\

\multicolumn{6}{l}{\textbf{Graph Classification}} \\
MUTAG & 73.2 $\pm$ 10.8 & 74.7 $\pm$ 10.1 & 71.5 $\pm$ 10.2 & \textbf{78.8 $\pm$ 13.5} & 73.2 $\pm$ 9.7 \\
PROTEINS & 71.5 $\pm$ 4.3 & 71.3 $\pm$ 3.5 & 70.1 $\pm$ 4.5 & 70.9 $\pm$ 4.5 & \textbf{73.7 $\pm$ 4.7} \\
\bottomrule
\end{tabular}
}
\end{table}
Results clearly illustrate that \alg is highly competitive across all tasks, except for link prediction on the Cora dataset, where it underperforms by only 0.3\% with comparable standard deviation, and on the graph classification task for the MUTAG dataset where GIN outperforms it, albeit with high standard deviation. Notably, all models exhibit high variance on graph classification tasks. These findings demonstrate that \alg offers superior or competitive performance and robustness compared to existing standard GNN frameworks.

\section{Ablation Study}\label{sec:ablation}
In this section, we evaluate key architectural and
hyperparameter choices underlying the \alg model. Specifically, we
assess the effectiveness of dynamically adjustable spline control-point
spacing, data-driven Gaussian-aligned initialisation, computational
efficiency of employing RBFs compared to
traditional B-splines, and sensitivity analyses concerning the spline
grid dimensions and number of control points. Each experiment is
motivated explicitly by the theoretical insights discussed previously
(Sections~\ref{sec:background},~\ref{sec:methodology}), and empirically
demonstrates the contributions to model performance and computational
efficiency.

\subsection{Control Point Initialisation and Spacing}\label{sec:ablation_control}
Fixed and evenly spaced control points restrict the model ability to adapt to varying data distribution. To address this limitation, we evaluate the effectiveness of dynamically adjustable knot spacing and Gaussian-based initialisation strategies. Table~\ref{tab:knot-init} presents the empirical results, clearly demonstrating that spline knots initialized from a Gaussian distribution (\textit{G}) with fully trainable spacing (\textit{T}) significantly outperform fixed, evenly spaced knots (\textit{E!T}).
\begin{table}[ht]
    \centering
    \caption{Ablation study evaluating knot initialisation and spacing strategies on the Cora dataset. Configuration labels: \textit{G} = Gaussian initialisation, \textit{E} = evenly spaced initialisation, \textit{T} = trainable spacing, \textit{!T} = fixed spacing. Mean accuracy over 10 runs is reported. Additional results are provided in Appendix~\ref{A:init}.}
    \label{tab:knot-init}
    \begin{tabular}{lc}
        \hline
        Configuration & Test Accuracy (\%)  \\
        \hline
        ET   & 77.8 $\pm$ 0.8  \\
        E!T  & 77.6 $\pm$ 1.3  \\
        G!T  & 79.8 $\pm$ 0.9  \\
        GT   & \textbf{82.0 $\pm$ 0.7} \\
        \hline
    \end{tabular}
\end{table}
The results in Table~\ref{tab:knot-init} clearly demonstrate that learnable knot placement with Gaussian initialisation (\textit{GT}) significantly improves accuracy compared to standard evenly spaced and fixed initialisation methods. These findings underscore the importance of adaptive spline flexibility for improved representational power and generalization performance.

\subsection{Activation Functions: B-splines vs. RBFs}\label{sec:ablation-rbf}
We investigate the use of RBFs, specifically Gaussian kernels, as computationally efficient alternatives to traditional B-spline activations, motivated by their simpler computation and flexibility. Table~\ref{tab:rbf-vs-splines} illustrates that RBF activations not only significantly outperform B-splines in terms of accuracy but also substantially reduce computational overhead, yielding an approximate fivefold reduction in training time per epoch. These findings suggest that Gaussian RBFs offer a compelling advantage for scaling the KANG architecture to larger graph datasets.
\begin{table}[ht]
    \centering
    \caption{Comparison of \alg performance using B-spline vs. RBF activations on the Cora dataset. Reported are mean test accuracy and average epoch training time over 10 runs. Additional results are provided in Appendix~\ref{A:rbf}.}
    \label{tab:rbf-vs-splines}
    \begin{tabular}{lcc}
        \hline
        Activation & Test Accuracy (\%) & Epoch Time (s) \\
        \hline
        B-splines   & 77.4 $\pm$ 0.9 & 0.063 $\pm$ 0.006 \\
        RBF        & \textbf{81.4 $\pm$ 0.8} & \textbf{0.012 $\pm$ 0.001} \\
        \hline
    \end{tabular}
\end{table}

\subsection{Sensitivity Analysis}
\begin{figure}[ht]
  \centering
  \begin{subfigure}[b]{0.47\textwidth}
      \centering
      \includegraphics[width=\textwidth]{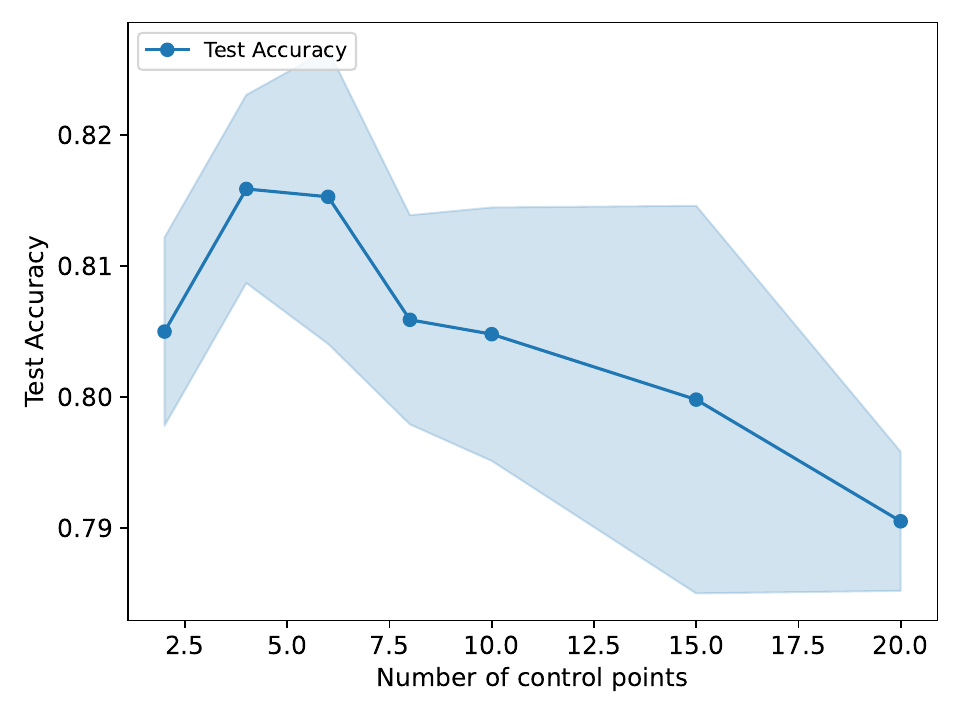}
      \caption{Test accuracy vs. number of control points.}
      \label{fig:accuracy}
  \end{subfigure}
  \hfill
  \begin{subfigure}[b]{0.47\textwidth}
      \centering
      \includegraphics[width=\textwidth]{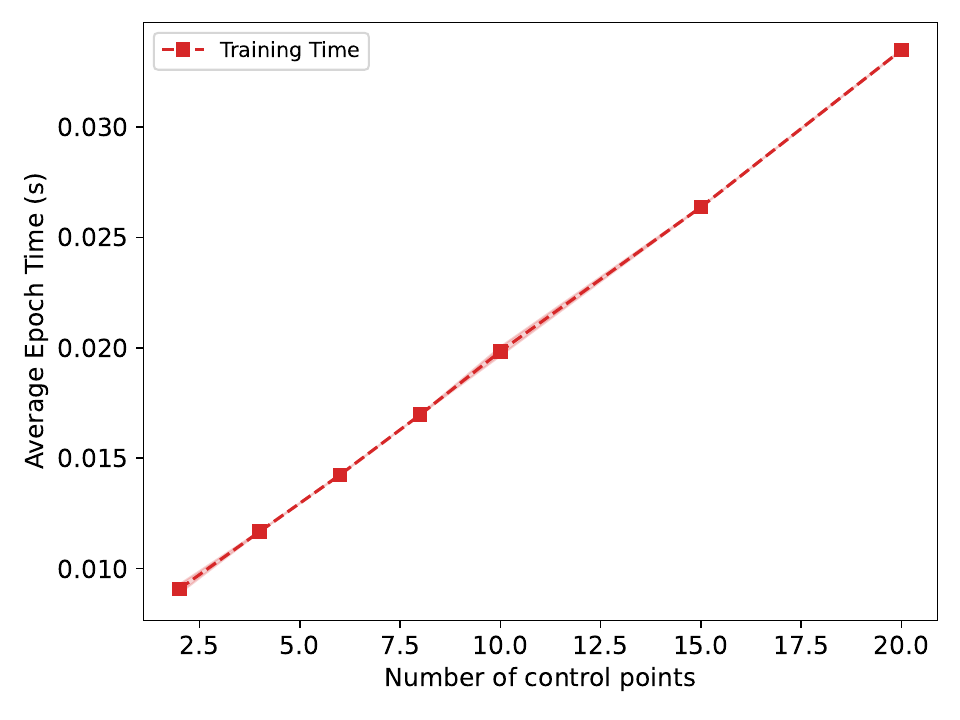}
      \caption{Epoch time vs. number of control points.}
      \label{fig:time}
  \end{subfigure}
  \caption{Sensitivity analysis on the Cora dataset for node classification. \textbf{Left}: Test accuracy decreases with more control points. \textbf{Right}: Epoch time increases with additional control points. Means and standard deviations computed over 10 runs per setting.}
  \label{fig:control_points}
\end{figure}

The number of control points (knots) is pivotal as it directly affects
model complexity and expressivity. Figure~\ref{fig:control_points}
presents a sensitivity analysis, showing an optimal choice at four
control points for node classification accuracy on the Cora dataset. The analysis further highlights a clear trade-off with computational efficiency, as shown by the near-linear increase in epoch time with additional control points. These findings underscore the importance of carefully selecting this hyperparameter to balance computational cost and predictive performance.

\section{Analysis}
In this section we will investigate the oversmoothing robustness of \alg. Together with its ability to scale to larger graphs, and how it is possible to visualise the learning process inside \alg.

\subsection{Oversmoothing}

Oversmoothing in GNNs refers to the phenomenon wherein node representations become nearly indistinguishable at deeper layers, leading to diminished discriminative power~\cite{chen2020simple,li2019deepgcns,huang2020tackling}. To mitigate this effect, we incorporate residual connections~\cite{li2019deepgcns,chen2020simple} into all tested architectures to assess their resilience by monitoring the Dirichlet energy $\mathcal{E}$ at increasing number of layers,
\begin{equation}\label{eq:de}
    \mathcal{E}(\mathbf{X})
    = \frac{1}{2 \lvert E \rvert} 
    \sum_{(u,v) \in E} \bigl\|\mathbf{x}_u - \mathbf{x}_v\bigr\|^2,
\end{equation}
where $\mathbf{X}$ denotes the node embedding matrix, $E$ is the set of edges, and $\lvert E\rvert$ is the total number of edges~\cite{rusch2023survey}.

\begin{figure}[ht]
    \centering
    \includegraphics[width=0.8\textwidth]{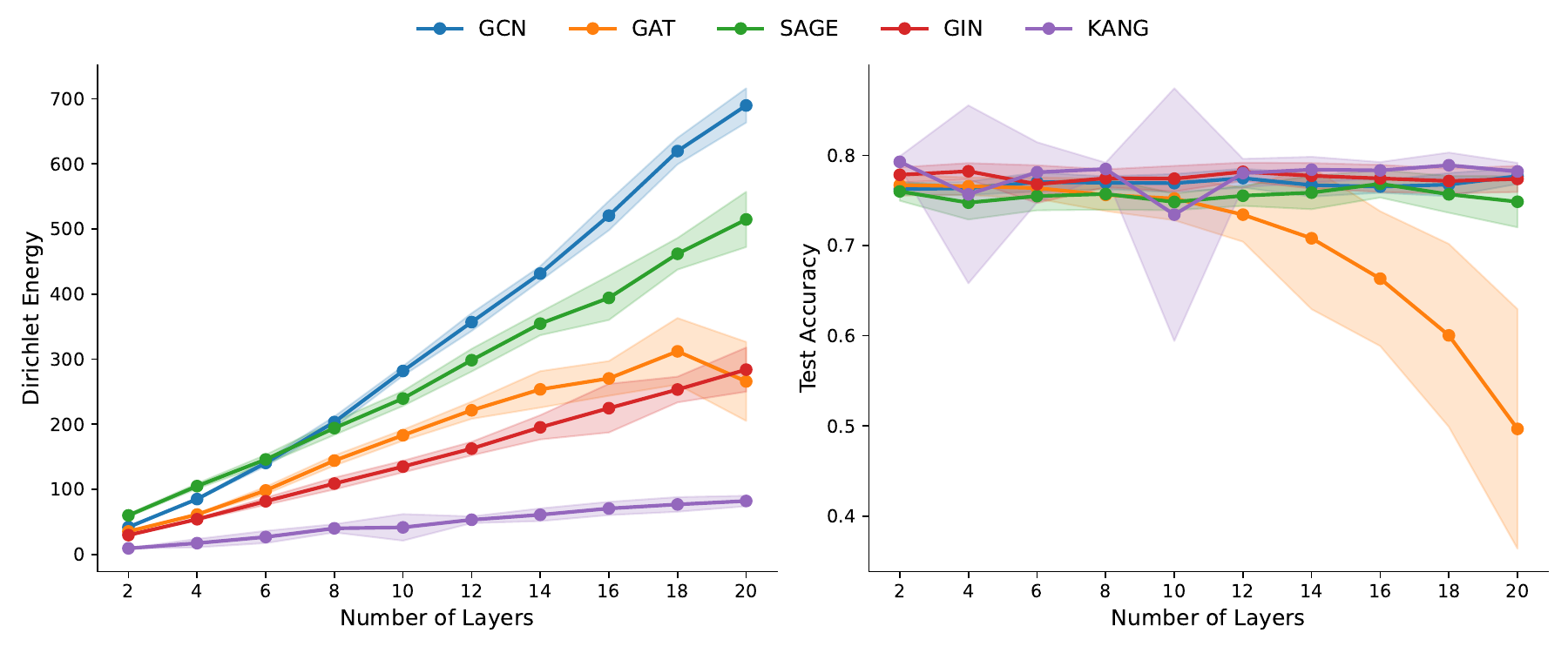}
    \caption{\textbf{Left:} Dirichlet Energy. \textbf{Right:} Test accuracy. Results obtained averaging over 10 runs with different random seed at each run. Additional results on the CiteSeer and PubMed datasets in Appendix~\ref{A:over}}
    \label{fig:oversmoothing}
\end{figure}

Figure~\ref{fig:oversmoothing} shows that, although \alg exhibits a lower $\mathcal{E}$ across all examined depths, it still maintains or even surpasses the performance of compared architectures. We attribute this apparent discrepancy to how \alg embeds nodes more compactly in feature space while preserving (or improving) linear separability, consistent with recent findings that smaller pairwise distances do not necessarily diminish classification accuracy~\cite{arroyo2025vanishing}. In contrast, the other architectures produce more dispersed embeddings that may not always yield better predictive performance.

Although it may seem counterintuitive, the increasing trend of Dirichlet Energy as the number of layers grows is expected when residuals are included in the architecture~\cite{zhou2021dirichlet}. Conversely, the trend is reversed in networks without residuals.

\subsection{Scalability}
\begin{figure}[ht]
    \centering
    \includegraphics[width=0.6\textwidth]{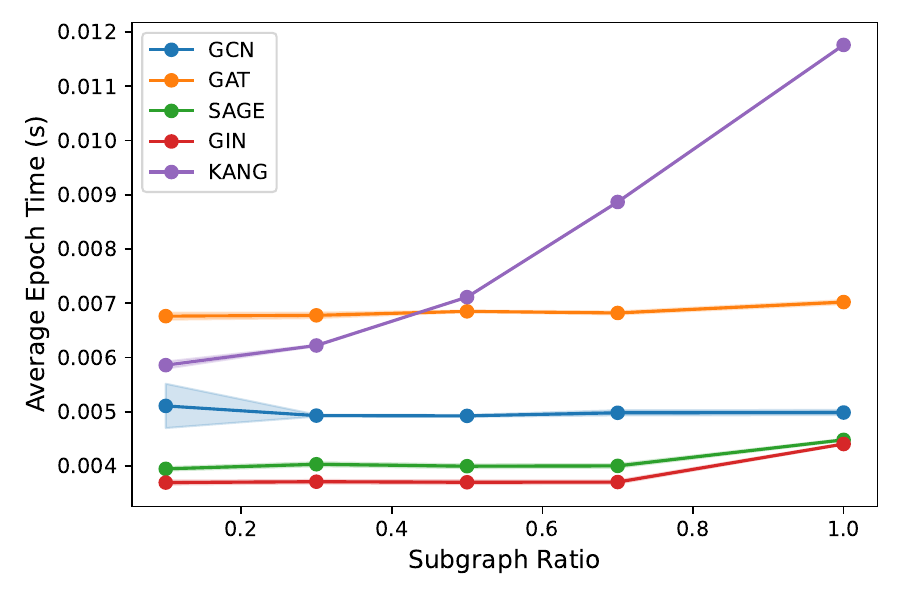}
    \caption{Computational scaling analysis across different subgraph ratios of the Cora dataset. Results obtained by averaging the results on 5 runs, with different random seeds.}
    \label{fig:scale}
\end{figure}

Figure~\ref{fig:scale} demonstrates the computational efficiency of \alg compared to baseline GNN architectures as the graph size increases. We evaluated performance by measuring average epoch training time across various subgraph ratios of the Cora dataset (additional results in Appendix~\ref{A:scale}). The results reveal that \alg maintains comparable computational efficiency to standard GNN architectures up to 50\% of the dataset dimension, but at larger graph sizes, its scaling behavior diverges from the baselines.

The observed behavior aligns with the theoretical expectations of KANs~\cite{liu2024kan}. Although \alg does not scale as efficiently as the compared architectures at larger graph sizes, it still exhibits linear scaling with respect to graph size (additional on the models dimesions can be found in Appendix~\ref{A:dim}).

\subsection{Spline Evolution Analysis}\label{sec:trans}
\begin{figure}[ht]
    \centering
    \begin{subfigure}[b]{0.49\textwidth}
        \centering
        \includegraphics[width=\textwidth]{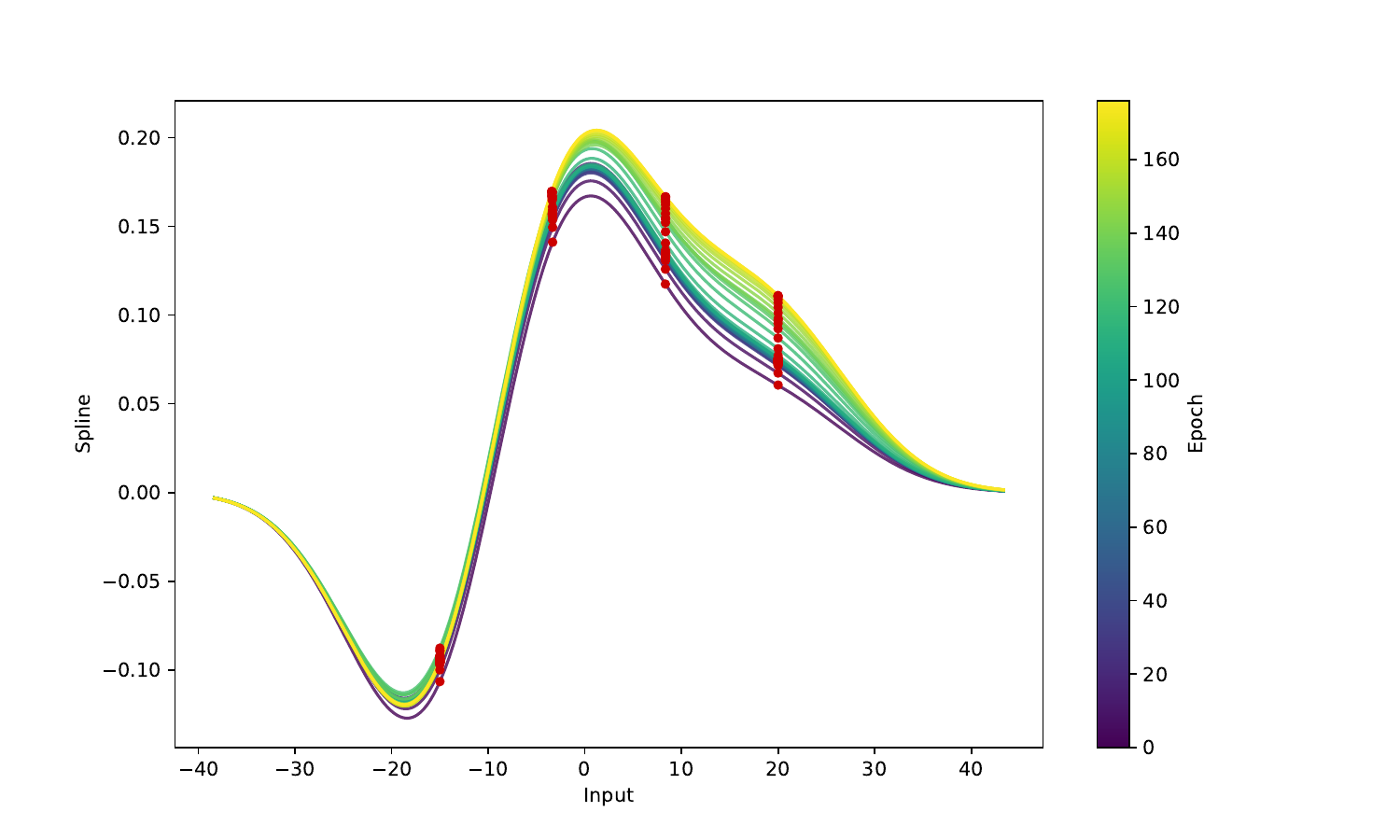}
        \label{fig:evo1}
    \end{subfigure}
    \begin{subfigure}[b]{0.49\textwidth}
        \centering
        \includegraphics[width=\textwidth]{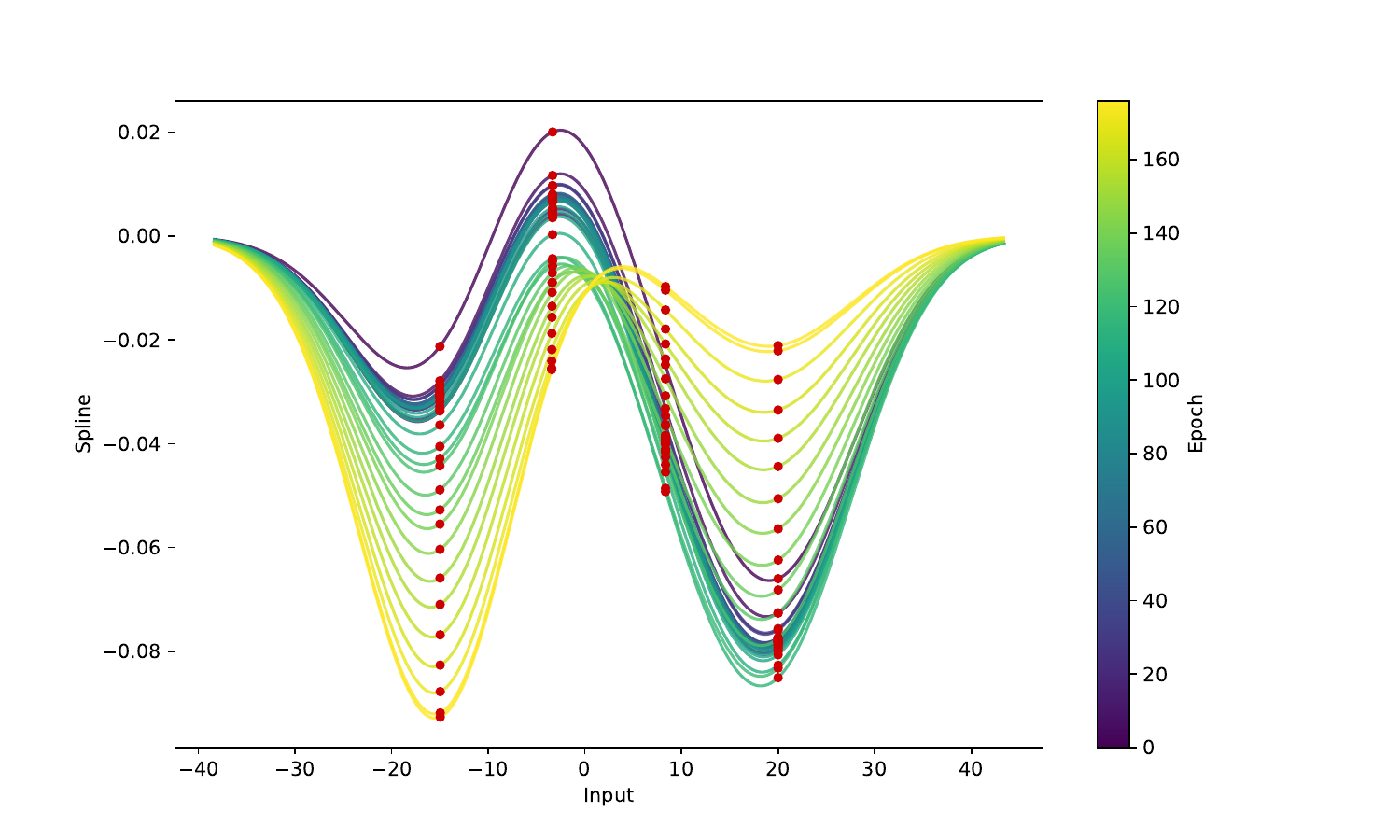}
        \label{fig:evo2}
    \end{subfigure}
    \caption{Temporal evolution of two representative spline activation functions across training epochs. The visualization reveals the progressive vertical and horizontal adaptation from near-linear behavior to complex non-linear transformations, demonstrating \alg's ability to automatically discover suitable activation functions for the given graph data.}
    \label{fig:evo}
  \end{figure}

The dynamic adaptation of spline activations during training provides valuable insights into \alg's learning process. Figure~\ref{fig:evo} illustrates the temporal evolution of a representative spline activation function across training epochs.

Initially, the spline exhibits a near-linear relationship, closely resembling a standard linear activation. As training progresses, we observe the emergence of increasingly complex, non-linear transformations that adapt to the specific characteristics of the input data distribution. This evolution demonstrates \alg's ability to dynamically adjust its activation landscape to capture intricate patterns in the graph structure. The final spline configuration displays pronounced non-linearities in specific regions of the input domain, indicating areas where the model has identified critical feature interactions. The splines at the end of the training are shown in Appendix~\ref{A:splines}.

The ability to analyze the learning characteristics of splines represents a significant step towards transparent GNNs. \alg's spline-based architecture allows direct inspection of the learned feature transformations, shedding light on the model's decision-making process. This transparency fosters the development of self-interpretable GNNs in the future.

\section{Conclusion}
In this paper, we introduce \alg, a novel graph neural network architecture that embeds the strengths of KAN's spline-based activation functions into graph learning. \alg leverages a data-driven initialisation strategy to create a flexible and expressive model that adapts to the specific characteristics of the input data, enabling it to capture complex relationships within graph structures. It achieves competitive performance in node classification, link prediction, and graph classification tasks on benchmark datasets against established architectures, while offering enhanced transparency through the visualization of learned spline activations. This transparency not only provides insights into the model's decision-making processes but also paves the way towards more self-interpretable graph learning models. However, the increased expressiveness comes with a trade-off: employing complex splines—with high degrees or a large number of control points—can significantly raise memory consumption, posing scalability challenges for large graphs. Ongoing efforts to improve the efficiency of learnable activation functions promise to address these challenges \cite{li2024kolmogorov}; this will foster the development of even more robust and scalable KAN-based GNNs suitable for real-world applications.



\bibliographystyle{splncs04}
\bibliography{ecml}

\newpage
\appendix
\section{Appendix}\label{sec:appendix}
\subsection{Implementation Details}\label{A:impl}

Our architecture is implemented in PyTorch~\cite{paszke2019pytorch} and PyTorch Geometric~\cite{fey2019fast}, leveraging GPU-accelerated tensor operations. We perform extensive hyperparameter optimization, as detailed in Section~\ref{sec:experiments}, ensuring reproducibility and state-of-the-art performance across benchmark datasets.

\subsection{Dataset Details}\label{A:data}
\begin{table}[ht]
  \centering
  \caption{Summary of benchmark datasets employed in experiments.}
  \label{tab:datasets}
  \begin{tabular}{lcccc}
  \toprule
  \textbf{Dataset} & \textbf{\#Nodes} & \textbf{\#Edges} & \textbf{\#Features} & \textbf{\#Classes} \\
  \midrule
  \multicolumn{5}{l}{\textbf{Node Classification and Link Prediction}} \\
  Cora & 2,708 & 10,556 & 1,433 & 7 \\
  PubMed & 3,327 & 9,104 & 3,703 & 6 \\
  CiteSeer & 19,717 & 88,648 & 500 & 3 \\
  \midrule
  \multicolumn{5}{l}{\textbf{Graph Classification}} \\
  MUTAG & 188 & \(\sim17.9\) & \(\sim\)39.6 & 2 \\
  PROTEINS & 1,113 & 39.1 & 145.6 & 2 \\
  \bottomrule
  \end{tabular}
  \end{table}

\subsection{\alg Hyperparameters}\label{A:hyper}

Table~\ref{tab:kang-hparams} presents the optimal hyperparameters identified for node classification (NC), link prediction (LP), and graph classification (GC). Here, \(h\) denotes the dimensionality of hidden layers, \(d\) is the dropout rate, \(\text{lr}\) and \(\text{wd}\) correspond to the learning rate and weight decay, respectively, and \(\text{grid\_min}\) and \(\text{grid\_max}\) define the range of spline control points. The column \(\text{knots}\) indicates the number of spline control points, while \(\text{layers}\) specifies the depth of the model.

\begin{table}[ht]
  \centering
  \caption{Optimal hyperparameters for \alg on the tasks of node classification (NC), link prediction (LP), and graph classification (GC).}
  \label{tab:kang-hparams}
  \begin{tabular}{lccccccccc}
    \toprule
    \textbf{Task} & \(\mathbf{h}\) & \(\mathbf{d}\) & \(\mathbf{lr}\) & \(\mathbf{wd}\) & \(\mathbf{grid\_min}\) & \(\mathbf{grid\_max}\) & \textbf{knots} & \textbf{layers} \\
    \midrule
    NC & 32 & 0.1 & 0.001 & 4e-4  & -15 & 20 & 4 & 2 \\
    LP & 8 & 0.1 & 0.008 & 5e-4 & -12  & 11 & 7 & 2 \\
    GC & 32 & 0.1 & 0.004 & 0.005 & -10 & 3  & 6 & 2 \\
    \bottomrule
  \end{tabular}
\end{table}

\subsection{Grid Range Sensitivity Analysis}\label{A:grid}
\begin{figure}[ht]
  \centering
  \includegraphics[width=0.7\textwidth]{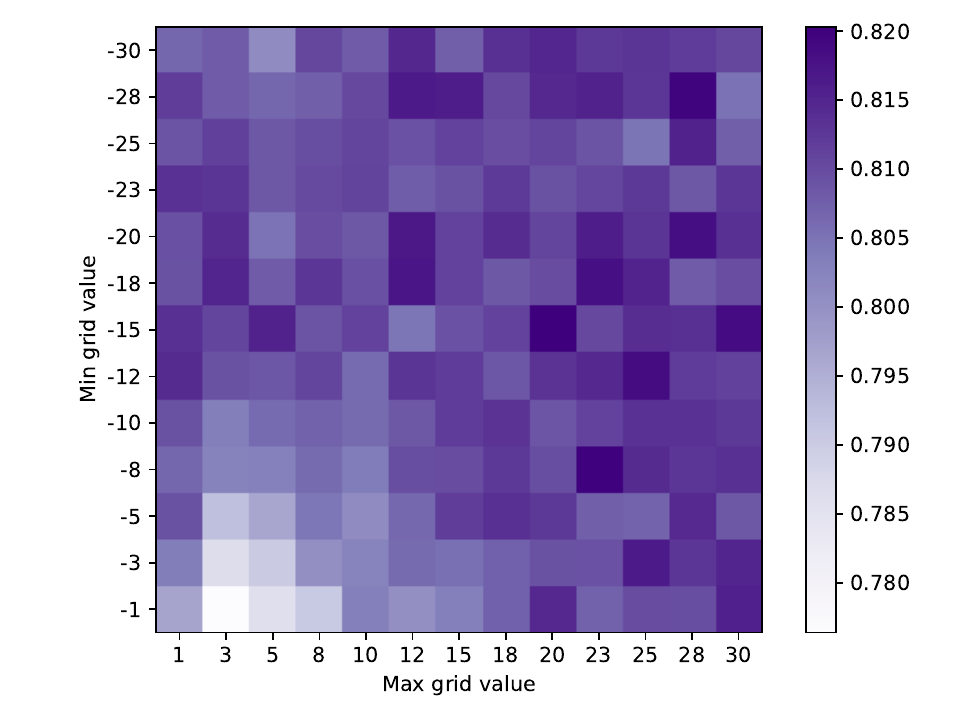}
  \caption{Sensitivity analysis illustrating model performance (test
    accuracy) as a function of minimum and maximum spline grid
    boundary values. Broader grid ranges improve model generalization
    and robustness.}
  \label{fig:grid-sensitivity}
\end{figure}

We systematically explore the sensitivity of model performance to spline
grid boundary hyperparameters (grid\_min, grid\_max).
Figure~\ref{fig:grid-sensitivity} demonstrates that broader grid ranges
consistently improve accuracy. Despite normalization of features,
extending the spline boundaries provides critical flexibility, enabling
the model to effectively handle sparse but informative regions of the
input feature distribution. This flexibility is particularly beneficial
for data characterized by heavy-tailed distributions or
outliers~\cite{bishop2006pattern,goodfellow2016deep}.

\subsection{Control Point Initialisation and Spacing}\label{A:init}
The initialisation and spacing of spline control points greatly affect model expressiveness. We compared four strategies: 
\begin{figure}[H]
  \centering
  \begin{subfigure}[b]{0.49\textwidth}
      \centering
      \includegraphics[width=\textwidth]{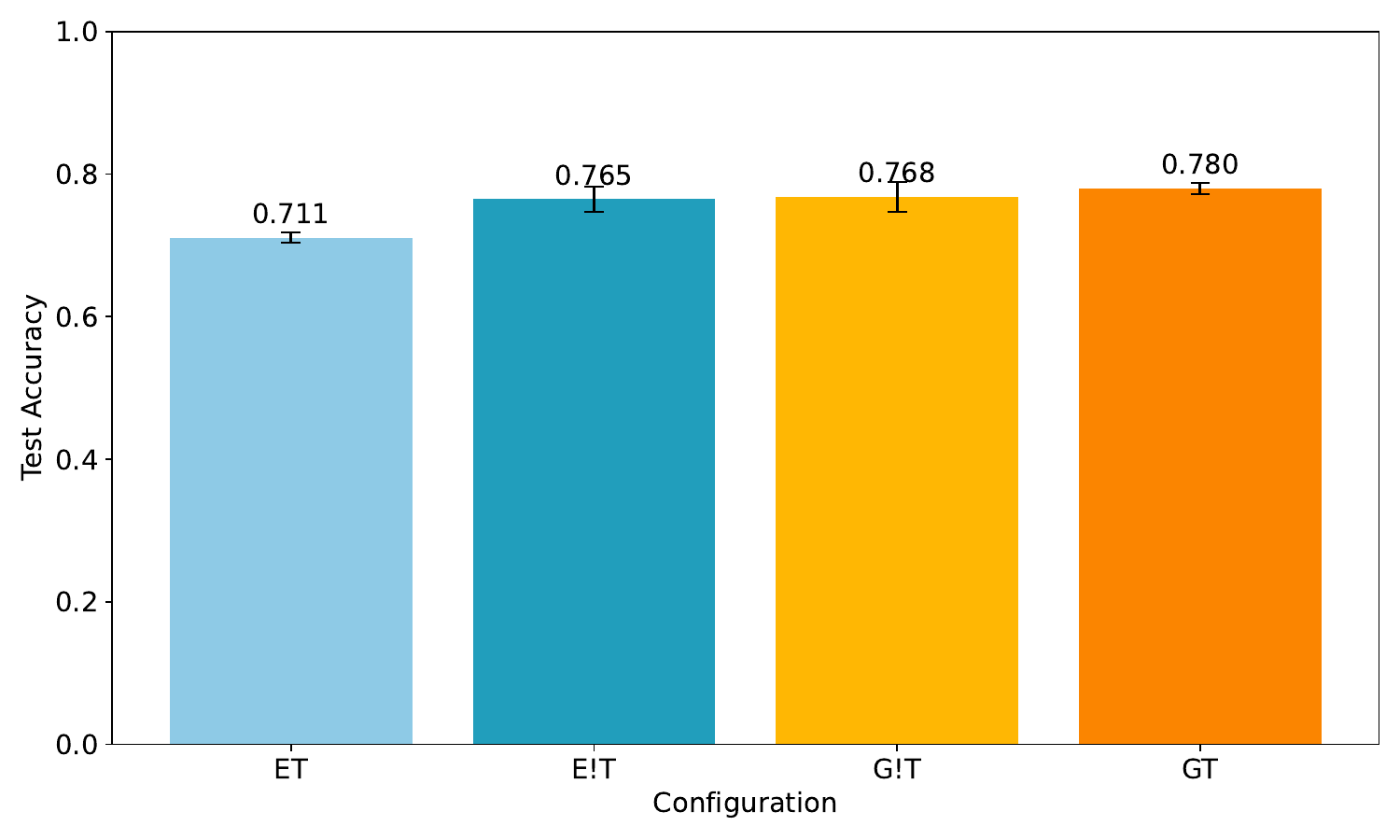}
      \caption{PubMed}
      \label{fig:control-PubMed}
  \end{subfigure}
  \hfill
  \begin{subfigure}[b]{0.49\textwidth}
      \centering
      \includegraphics[width=\textwidth]{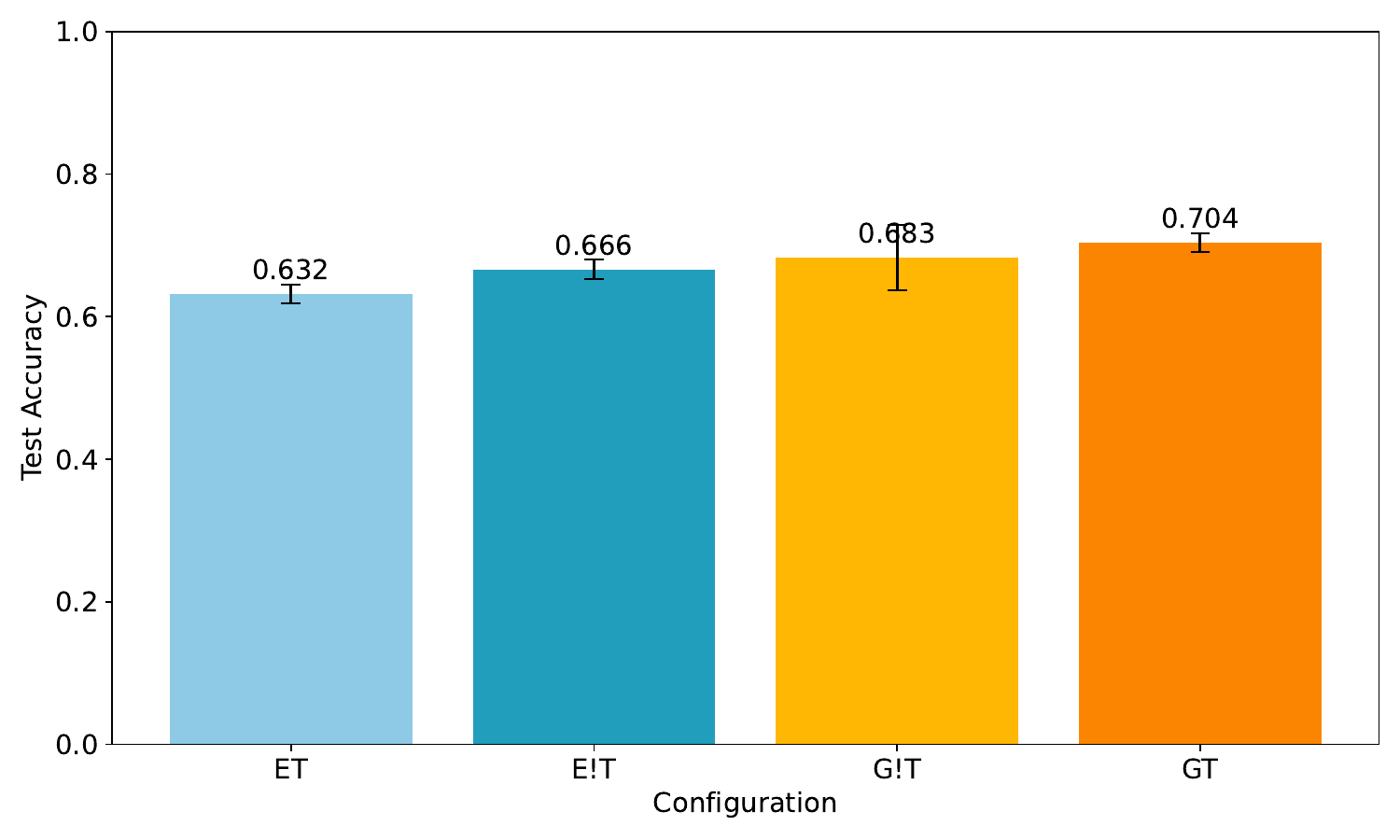}
      \caption{CiteSeer}
      \label{fig:control-CiteSeer}
  \end{subfigure}
  \caption{Ablation study results on different datasets.}
  \label{fig:control-all}
\end{figure}

\subsection{Activation Functions: B-splines vs. RBFs}\label{A:rbf}

\begin{figure}[H]
  \centering
  \begin{subfigure}[b]{0.49\textwidth}
      \centering
      \includegraphics[width=\textwidth]{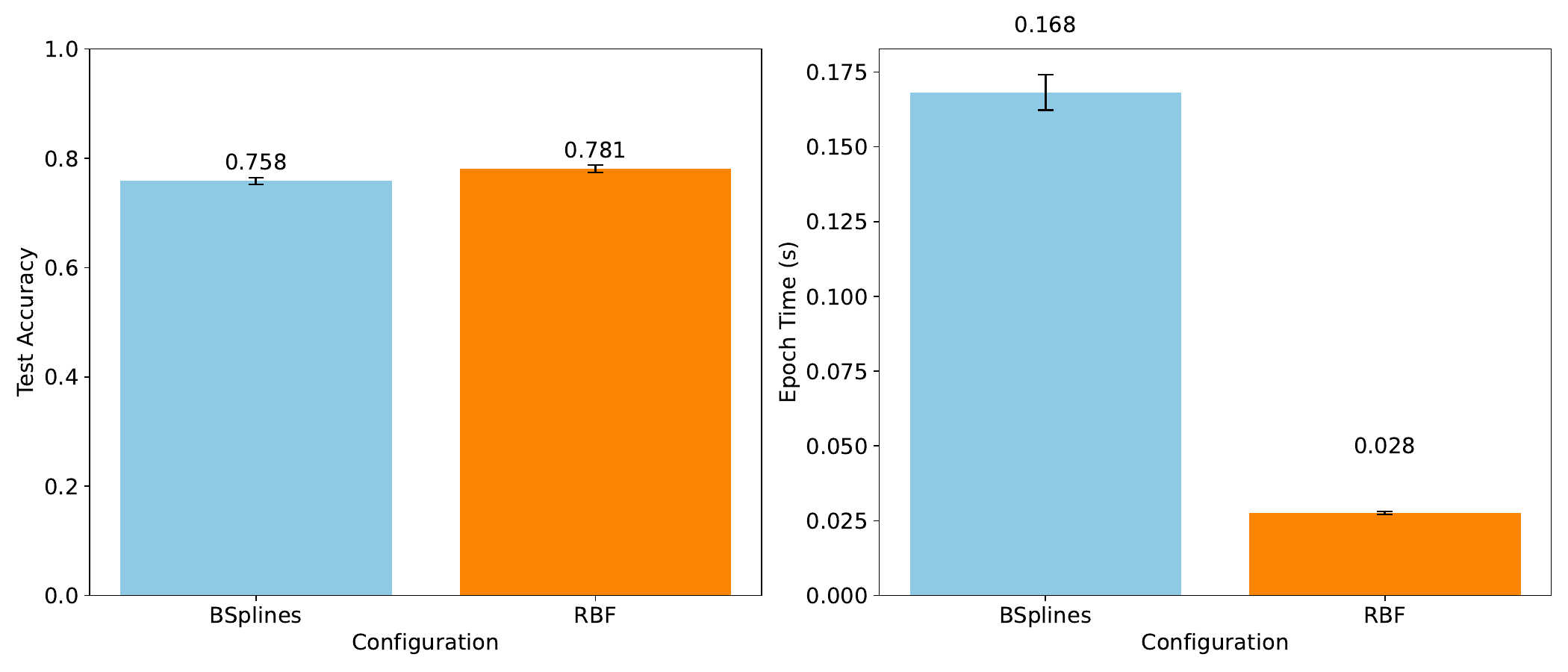}
      \caption{PubMed}
      \label{fig:act-PubMed}
  \end{subfigure}
  \hfill
  \begin{subfigure}[b]{0.49\textwidth}
      \centering
      \includegraphics[width=\textwidth]{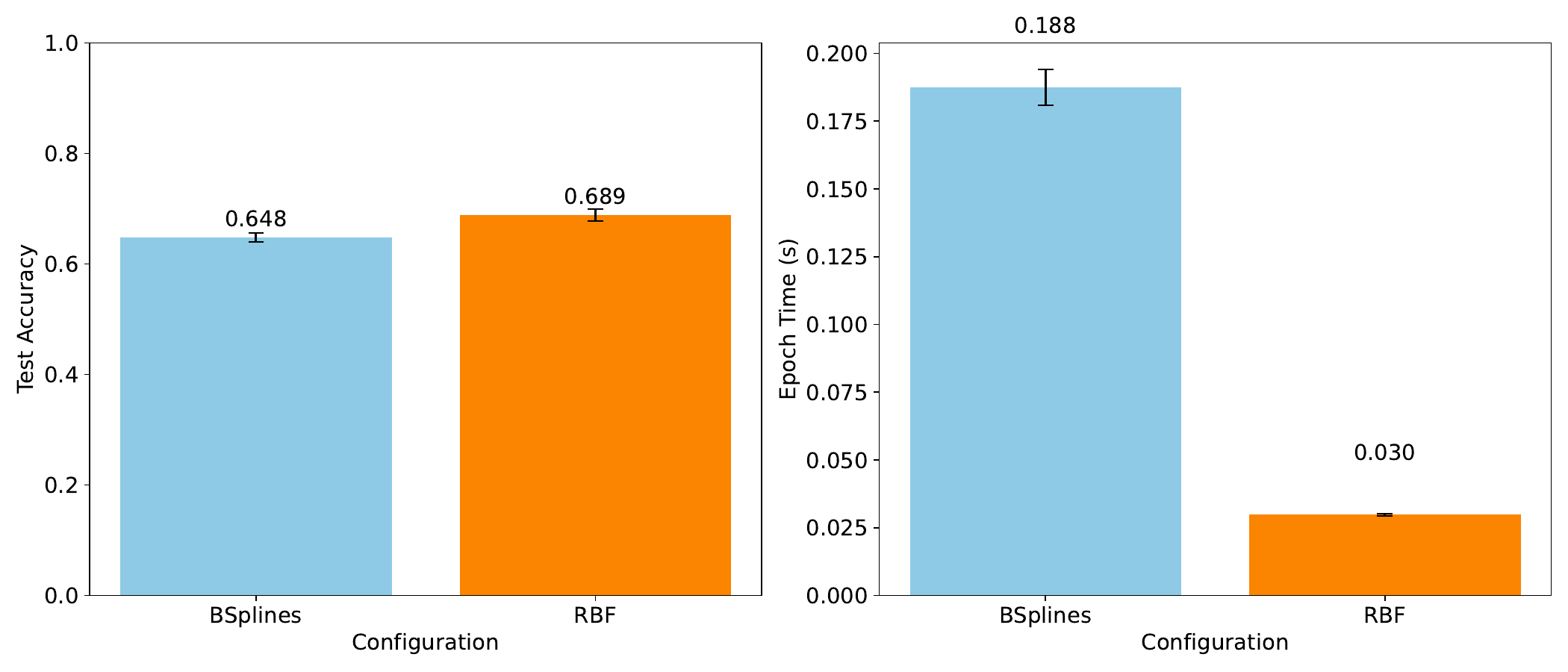}
      \caption{CiteSeer}
      \label{fig:act-CiteSeer}
  \end{subfigure}
  \caption{Ablation study results on different using B-splines or RBF as learnable activatio functions.}
  \label{fig:act-all}
\end{figure}

\subsection{Oversmoothing}\label{A:over}
\begin{figure}[H]
  \centering
  \begin{subfigure}[b]{0.49\textwidth}
      \centering
      \includegraphics[width=\textwidth]{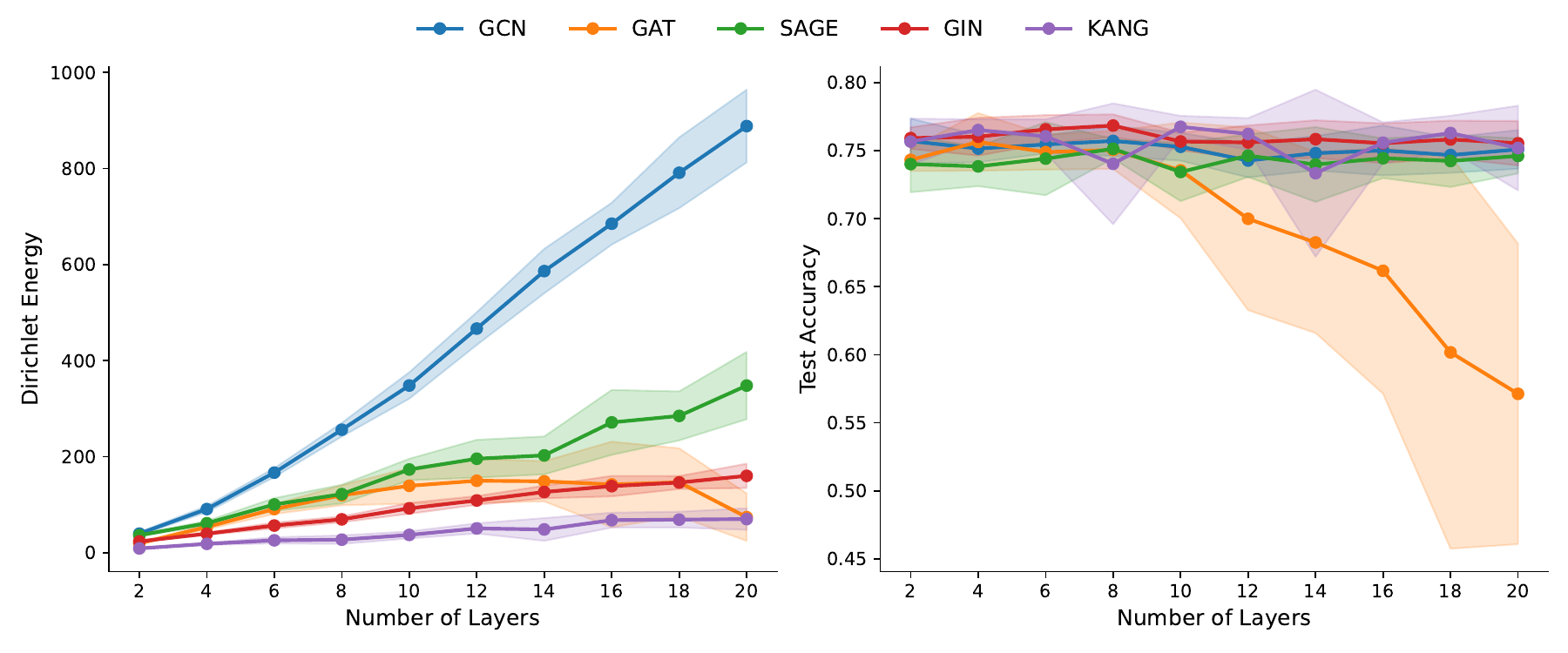}
      \caption{PubMed}
      \label{fig:over-PubMed}
  \end{subfigure}
  \hfill
  \begin{subfigure}[b]{0.49\textwidth}
      \centering
      \includegraphics[width=\textwidth]{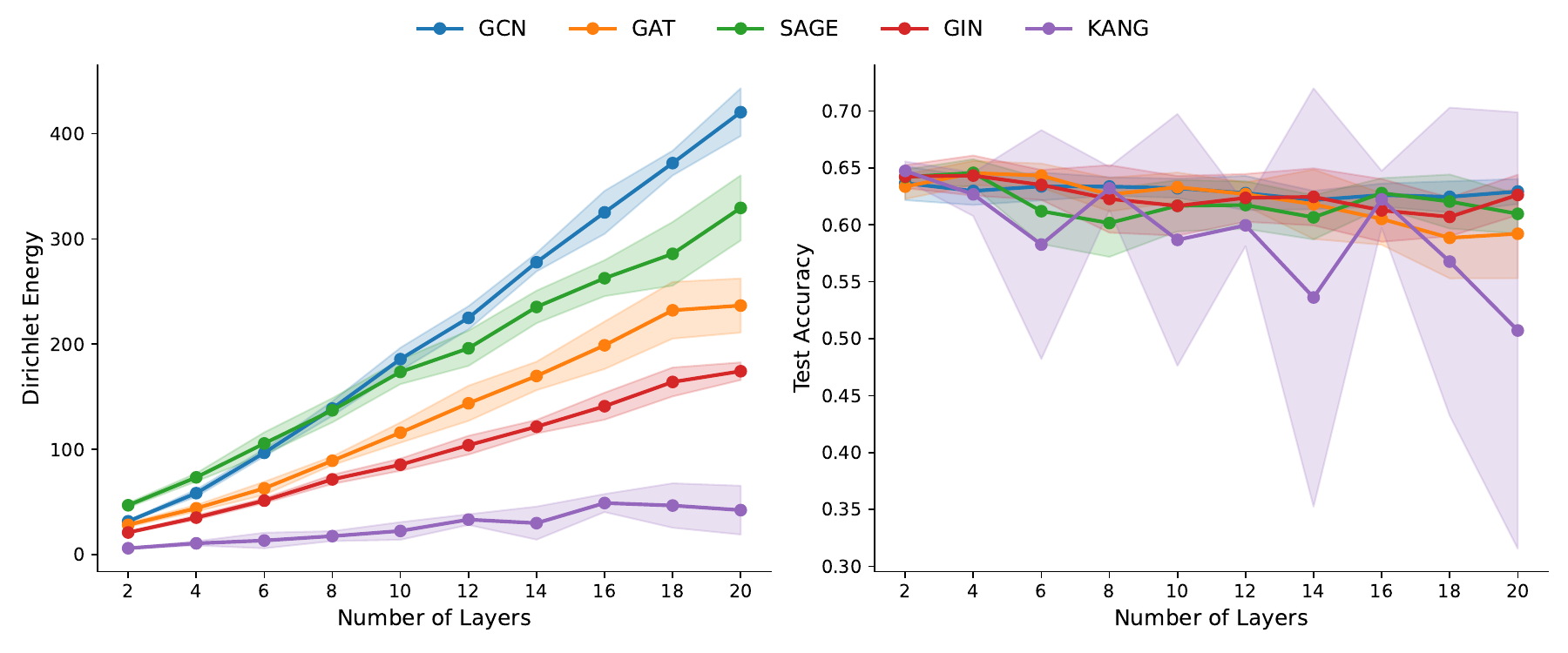}
      \caption{CiteSeer}
      \label{fig:over-CiteSeer}
  \end{subfigure}
  \caption{Oversmoothing analysis.}
  \label{fig:over-all}
\end{figure}

\subsection{Scalability}\label{A:scale}

\begin{figure}[H]
  \centering
  \begin{subfigure}[b]{0.49\textwidth}
      \centering
      \includegraphics[width=\textwidth]{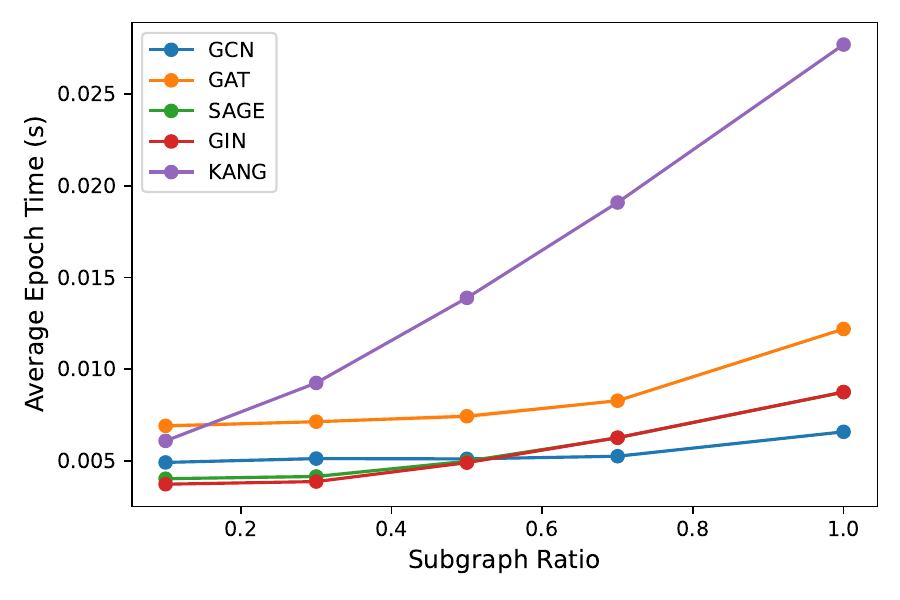}
      \caption{PubMed}
      \label{fig:scale-PubMed}
  \end{subfigure}
  \hfill
  \begin{subfigure}[b]{0.49\textwidth}
      \centering
      \includegraphics[width=\textwidth]{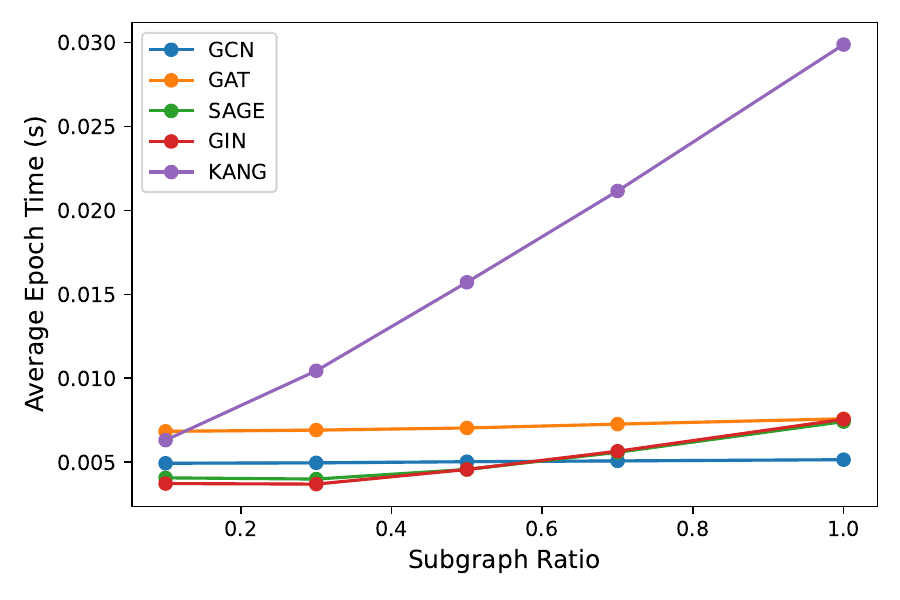}
      \caption{CiteSeer}
      \label{fig:scale-CiteSeer}
  \end{subfigure}
  \caption{Scalability analysis.}
  \label{fig:scale-all}
\end{figure}

\subsection{Model Dimensions}\label{A:dim}

The complexity of a deep learning model can be assessed through trainable parameters and floating-point operations (FLOPs). Trainable parameters refer to the model's weights, which influence its capacity but do not directly reflect computational cost. Instead, FLOPs, which quantify the total arithmetic operations in a forward pass, provide a more reliable measure of complexity. As noted in prior work, FLOPs are the key metric when evaluating computational efficiency~\cite{thompson2021deep}.

\begin{table}[H]
  \centering
  \caption{Number of trainable parameters and FLOPs for the compared GNNs, measured on the Cora dataset.}
  \label{tab:model_dimensions}
  \begin{tabular}{lcc}
    \toprule
    \textbf{GNN} & \textbf{Trainable Parameters} & \textbf{FLOPs} \\
    \midrule
    GCN & 96,647 & 525,517,440 \\
    GATv2 & 192,839 & 1,051,098,624 \\
    GraphSAGE & 192,455 & 1,040,911,872 \\
    GIN & 96,647 & 530,123,496 \\
    \alg & 238,597 & 1,283,684,072 \\
    \bottomrule
  \end{tabular}
\end{table}

\subsection{Splines}\label{A:splines}
\begin{figure}[H]
  \centering
  \begin{subfigure}[b]{0.32\textwidth}
      \centering
      \includegraphics[width=\textwidth]{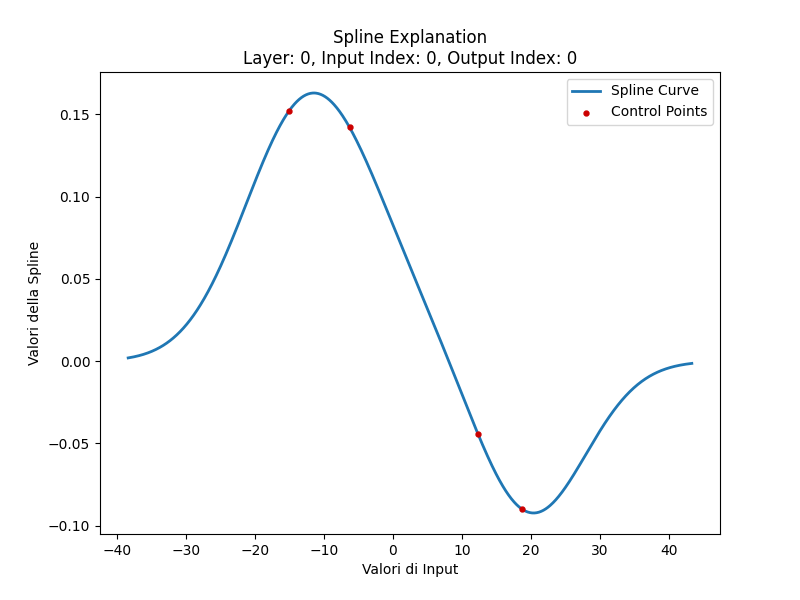}
      \label{fig:spline_0}
  \end{subfigure}
  \hfill
  \begin{subfigure}[b]{0.32\textwidth}
    \centering
    \includegraphics[width=\textwidth]{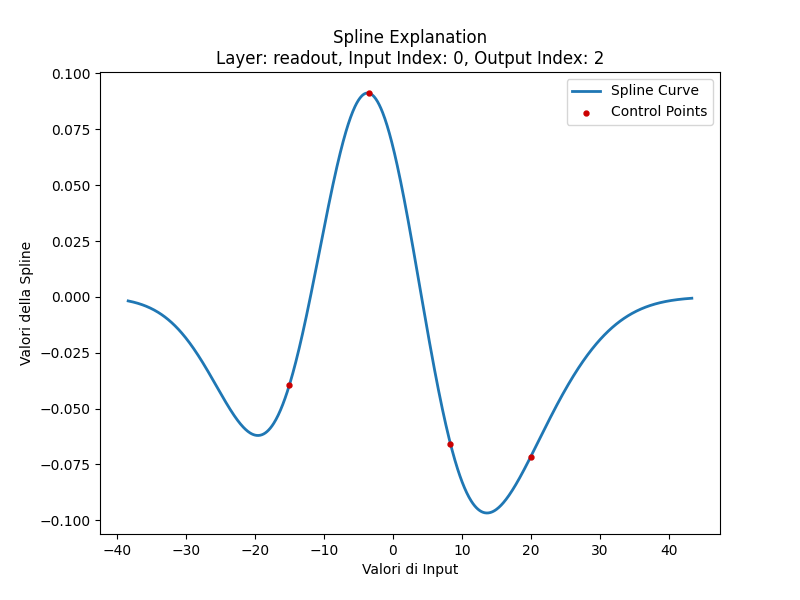}
    \label{fig:spline_1}
\end{subfigure}
\hfill
  \begin{subfigure}[b]{0.32\textwidth}
      \centering
      \includegraphics[width=\textwidth]{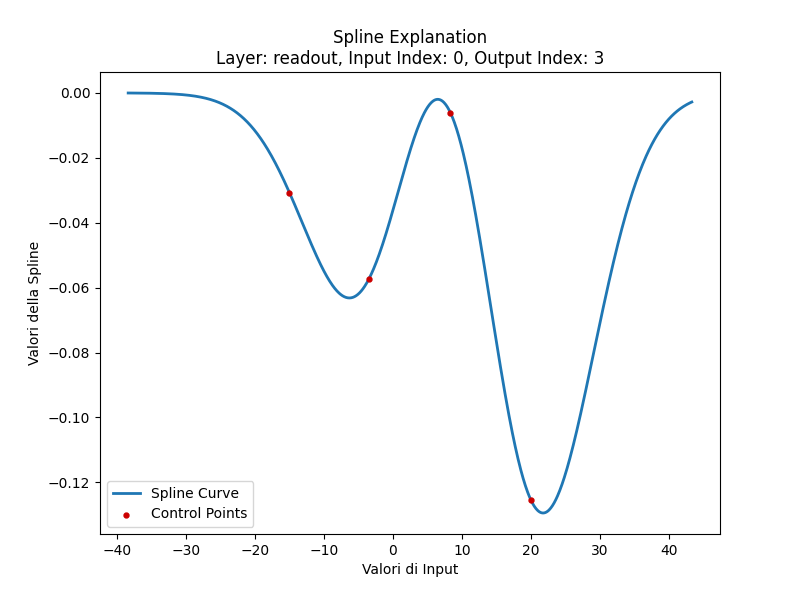}
      \label{fig:spline_2}
  \end{subfigure}
  \caption{Final learned splines for three neurons of \alg.}
  \label{fig:spline-all}
\end{figure}

\end{document}